\documentclass[10pt,twocolumn,letterpaper]{article}

\usepackage[pagenumbers]{cvpr} 

\usepackage{booktabs}
\usepackage{multirow}
\usepackage{balance}
\renewcommand{\paragraph}[1]{\vspace{1mm}\noindent\textbf{#1}}

\definecolor{cvprblue}{rgb}{0.21,0.49,0.74}
\usepackage[pagebackref,breaklinks,colorlinks,allcolors=cvprblue]{hyperref}
\def\paperID{23} 
\def\confName{CVPR}
\def\confYear{2025}

\title{Investigating Mechanisms for In-Context Vision Language Binding}

\author{
Darshana Saravanan\hspace{1cm}
Makarand Tapaswi\hspace{1cm}
Vineet Gandhi\vspace{1mm}\\
{\normalsize CVIT, IIIT Hyderabad, India}\vspace{1mm}\\
}

\begin{document}
\maketitle
\begin{abstract}
To understand a prompt, Vision-Language models (VLMs) must perceive the image, comprehend the text, and build associations within and across both modalities. For instance, given an `image of a red toy car', the model should associate this image to phrases like `car', `red toy', `red object', \etc.
\citet{feng2024how}~propose the Binding ID mechanism in LLMs, suggesting that the entity and its corresponding attribute tokens share a Binding ID in the model activations.
We investigate this for image-text binding in VLMs using a synthetic dataset and task that requires models to associate 3D objects in an image with their descriptions in the text.
Our experiments demonstrate that VLMs assign a distinct Binding ID to an object's image tokens and its textual references, enabling in-context association.
\end{abstract}
    
\section{Introduction}
\label{sec:intro}
\begin{figure}
\centering
\includegraphics[width=.4\textwidth]{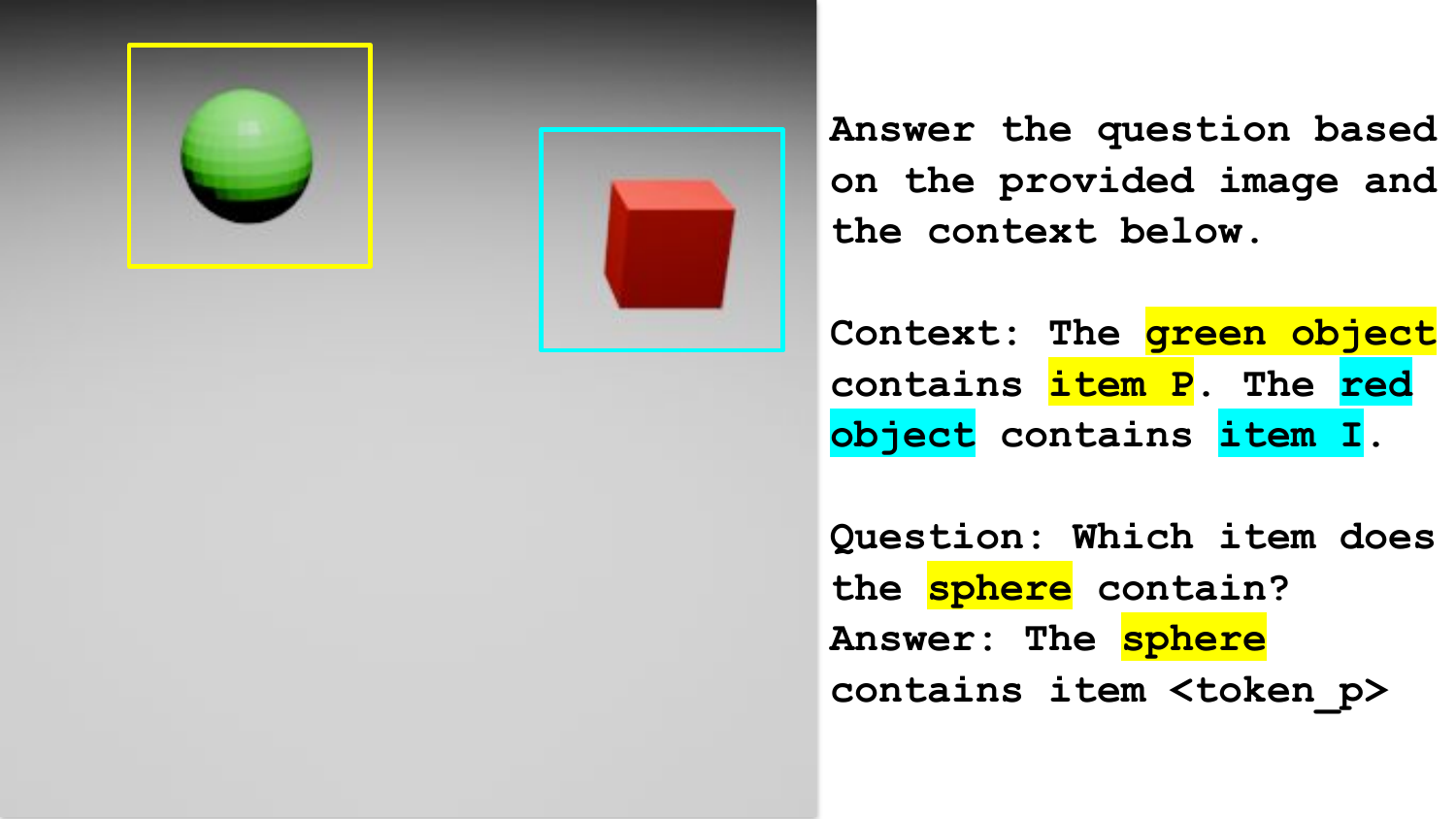}

\vspace{-1mm}
\caption{\textbf{Shapes Task.}
Given an image with two 3D objects and a text description (context), the model needs to comprehend the question and identify the correct item (\texttt{token\_p}) contained in the queried object.
Image and text tokens highlighted with the same color are expected to contain the same binding IDs, allowing the model to predict the correct answer.}
\label{fig:shapes_task}
\vspace{-3mm}
\end{figure}

As Vision-Language models (VLMs) like Gemini~\cite{team2024gemini} and GPT-4o~\cite{hurst2024gpt} become ubiquitous, it is crucial to understand how they function to determine why they respond the way they do, especially in safety-critical applications. A fundamental ability of VLMs is to associate information across an image and text to reason about a query.
For example, given an \textit{image of a furniture store that has a chair with a yellow tag} and the caption \textit{All furniture with a yellow tag have a 30\% discount}, a VLM should be able to infer that the chair has a discounted selling price. 
Our goal is to study this ability to \textit{bind} objects in an image to information in text.
To this end, we propose the \textit{Shapes} task, a controlled synthetic task that requires models to associate 3D objects in an image with their references in the text.
In \cref{fig:shapes_task}, the image contains two 3D objects: a \textit{green sphere} and a \textit{red cube}. 
The green sphere is referred to as the \textit{green object} in the context. 
So, to answer the question \textit{`What does the sphere contain?'}, the model needs to internally learn that the sphere corresponds to the phrase green object: \textit{is(`green sphere patches', `green object')},
and that this green object contains item P: \textit{contains(`green object', `item P')}.

The binding ID mechanism proposed in~\cite{feng2024how} suggests that LLMs' internal activations represent binding information by attaching binding ID vectors to the corresponding entities and attributes. We investigate whether VLMs use a similar mechanism to represent associations between image tokens and text tokens.
We study the most commonly used
VLM architecture that consists of a visual encoder, a multi-modal projector and a language model.

VLMs and LLMs have some key differences that necessitate careful experimentation.
(i)~Text tokens have fixed embeddings, while concepts in an image (objects, colors, textures, \etc.) do not have fixed embeddings; they are represented in the patch tokens obtained from the vision encoder.
(ii)~Recent powerful VLMs like LLaVA-OneVision~\cite{li2024llava}, Molmo~\cite{deitke2024molmo}, and Qwen2-VL~\cite{wang2024qwen2} utilize an image encoder that converts the input image into a set of multiscale, multi-crop images and independently maps each of these images into a set of vision tokens. This leads to multiple sets of tokens for the same visual concept. 
We adapt the causal mediation based experiments from~\cite{feng2024how} to account for these differences and make the following observations:
(i)~Image tokens corresponding to the location of the visual concept represent information related to that concept. This is applicable even when there are multiple tokens corresponding to multiple crops from the same image.
(ii)~VLMs implement the binding ID mechanism. There are binding ID vectors that associate the image tokens corresponding to a visual object and its references in the text tokens.

\section{Task Definition  and Notations}
\label{sec:task_definition}

\paragraph{Shapes task.}
This task consists of images with two 3D objects ($O_0$, $O_1$) with distinct shapes and colors.
The context refers to both objects using their color ($C_0$, $C_1$) and assigns a unique item ($I_0$, $I_1$).
We use the notation ${c = ctxt(O_0\leftrightarrow C_0\leftrightarrow I_0, O_1\leftrightarrow C_1\leftrightarrow I_1)}$ to denote a context where object $O_0$ of the color $C_0$ contains item $I_0$ and object $O_1$ of the color $C_1$ contains item $I_1$. In \cref{fig:shapes_task}, $O_0$ and $O_1$ correspond to the \textit{green sphere} and \textit{red cube} patches in the image, $C_0$ and $C_1$ correspond to \textit{green} and \textit{red} in the text and $I_0$ and $I_1$ correspond to \textit{item P} and \textit{item I} in text respectively. The question refers to one of the objects using its shape and queries the item assigned to it. Note that `\textit{item P/I}' are randomly chosen uppercase English letters with no inherent meaning.

We generate the images using Blender~\cite{blender}. We consider four choices for the shape (cone, cube, cylinder and sphere) and six choices for the color (red, blue, green, yellow, cyan and purple). The objects occupy a fixed number of patches and are located in fixed positions.

\paragraph{Notation.} 
Let $\Phi_v(\cdot)$ denote the vision encoder and $g(\cdot)$ denote the multi-modal projector.
For an image $X_v$, the patch embeddings are $t_v = g(\Phi_v(X_v))$.
%
Now, let $t_c$ denote the prompt tokens, comprising image tokens $t_v$ and text tokens up to the context's end, just before the question.
Let the LLM have $L$ transformer layers and $D$-dimensional activation space.
For every token position $p$, $Z_p\in R^{L \times D}$ is the stacked set of residual stream activations.
The activations at the object, color, and item positions are denoted as $Z_{O_k}$, $Z_{C_k}$ and $Z_{I_k}$ respectively where $k \in \{0,1\}$.

\section{Do Binding IDs Occur in VLMs?}
\label{sec: binding_ids_exist}

\paragraph{Binding Id Mechanism.}
\citet{feng2024how}~suggest that LLMs associate concepts
through binding ID vectors in their activations.
Specifically, the activations of an LLM can be decomposed into vectors that encode the concept and those that encode the binding information. Each binding ID consists of similar vector pairs in a subspace, with associated concepts sharing one vector from the same ID.
Extending this, we describe our hypothesis for the existence of binding IDs in VLMs using the Shapes task below:
\begin{itemize}
\item Consider 3D objects as visual entities and their colors and items mentioned in the text as their attributes. For the $k^\text{th}$ visual entity-attributes tuple $(I_k, C_k, O_k)$, the model represents binding vectors in its activations in an abstract form, independent of any particular object, color, or item.
\item For object patch tokens, the activations $Z_{O_k}$ can be decomposed as $Z_{O_k} = f_O(O_k) + b_O(k) $.
Similarly, $Z_{C_k} = f_C(C_k) + b_C(k)$ and $Z_{I_k} = f_I(I_k) + b_I(k)$.
Here $f_O(O_k)$, $f_C(C_k)$, $f_I(I_k)$ are the content vectors and the set of binding vectors $(b_O(k), b_C(k), b_I(k))$ form the binding ID for the $k^\text{th}$ tuple.
\item To answer the question about an object, the model selects the item that shares the same binding ID.
\end{itemize}

Note that, since binding IDs are independent of the entity/attribute, we can manipulate the associations built by the model by exchanging the binding IDs in the activations as $\hat{Z}_{O_k} := Z_{O_k} - b_O(k) + b_O(k')$ where $k \neq k'$.
In the following sections, we assert the existence of the binding ID mechanism by establishing two of its properties: Factorizability (\cref{subsubsec: factorizability}) and Position independence (\cref{subsubsec: position independence}). Then, we exchange the associations built by the model using Mean interventions (\cref{subsubsec: mean interventions}).

\subsection{Factorizability}
\label{subsubsec: factorizability}

\begin{figure}
\centering
\includegraphics[width=0.99\linewidth]{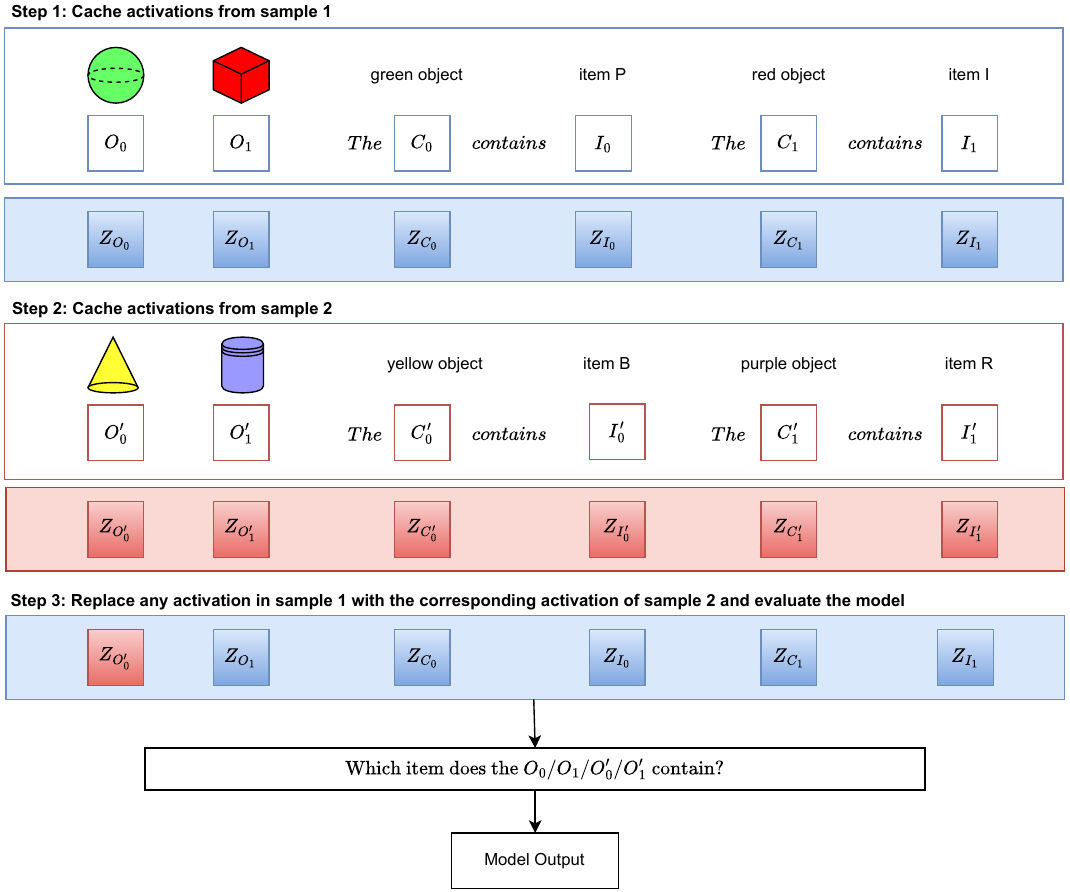}
\vspace{-2mm}
\caption{Causal intervention. In steps 1 and 2, activations from the first and second samples are saved. In step 3, object/color/item activations in the first sample are replaced with those from the second. This new activation stack is frozen, and the model is queried with all four objects to observe the change in predictions.}
\label{fig:causal_intervene}
\vspace{-3mm}
\end{figure}

\cref{fig:causal_intervene} shows two samples from the Shapes task with the contexts $c = ctxt(O_0\leftrightarrow C_0\leftrightarrow I_0, O_1\leftrightarrow C_1\leftrightarrow I_1)$  and $c' = ctxt(O'_0\leftrightarrow C'_0\leftrightarrow I'_0, O'_1\leftrightarrow C'_1\leftrightarrow I'_1)$. 

The Binding ID mechanism assumes that the information linking a concept to its attributes is stored locally within the activations at its token positions and is independent of the specific concept itself. This implies that the activations of the \textit{sphere} ($O_0$)  in the first sample and the \textit{cone} ($O'_0$) in the second sample should contain the same binding vector $b_O(0)$ as they both correspond to the $0^\text{th}$ visual entity-attributes tuple in their respective samples. 
Replacing $Z_{O_0}$ with $Z_{O'_0}$ should now bind the \textit{cone} with the text tokens \textit{green object} and \textit{item P}. We demonstrate this using causal interventions~\cite{vig2020investigating} on the activations as described below.


\begin{itemize}
\item Cache all activations $Z_c$ from the model run on $c$.
\item Cache activations $Z_{O'_0}$ and $Z_{O'_1}$ from the model run on $c'$.
\item Construct a new stack of activations $Z_c^{*}$ by modifying $Z_c$ such that $Z_{O_k}$ is replaced with $Z_{O'_k}$ for any $k \in \{0,1\}$.
\item Re-evaluate the model by probing what item each shape ($O_0, O_1, O'_0, O'_1$) contains by freezing the activation cache as $Z_c^{*}$. We expect the model to now associate $O'_k$ with $I_k$ since both $Z_{O_k}$ and $Z_{O'_k}$ contain the same binding ID vector $b_O(k)$.
\end{itemize}

\begin{figure}[t]
\centering
\begin{subfigure}[b]{\linewidth}
    \centering
    \includegraphics[width=\textwidth]{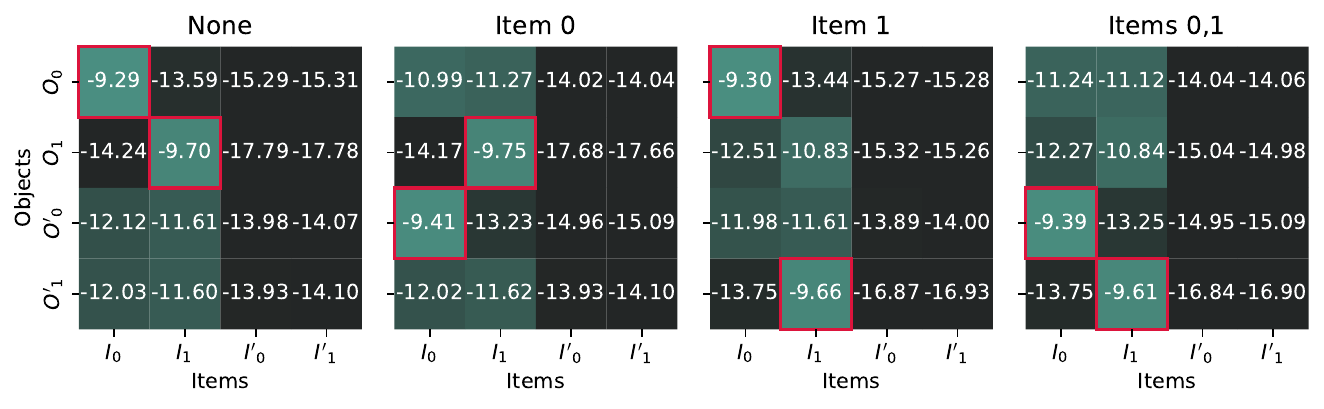}\vspace{-0.5em}
    \caption{Object activation replacements}
    \label{fig:fact_object}
\end{subfigure}
\hfill
\begin{subfigure}[b]{\linewidth}
    \centering
    \includegraphics[width=\textwidth]{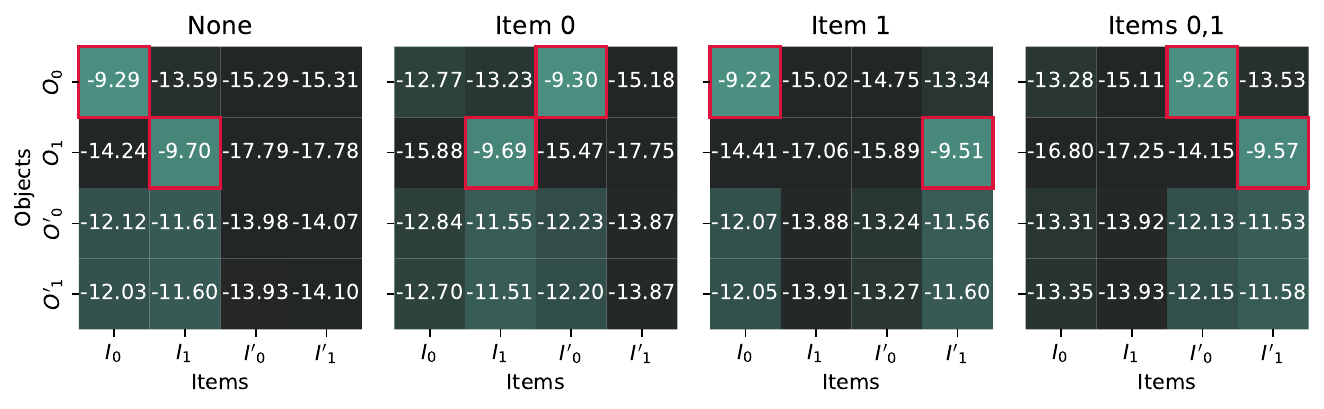}\vspace{-0.5em}
    \caption{Item activation replacements}
    \label{fig:fact_item}
\end{subfigure}
\hfill
\begin{subfigure}[b]{\linewidth}
    \centering
    \includegraphics[width=\textwidth]{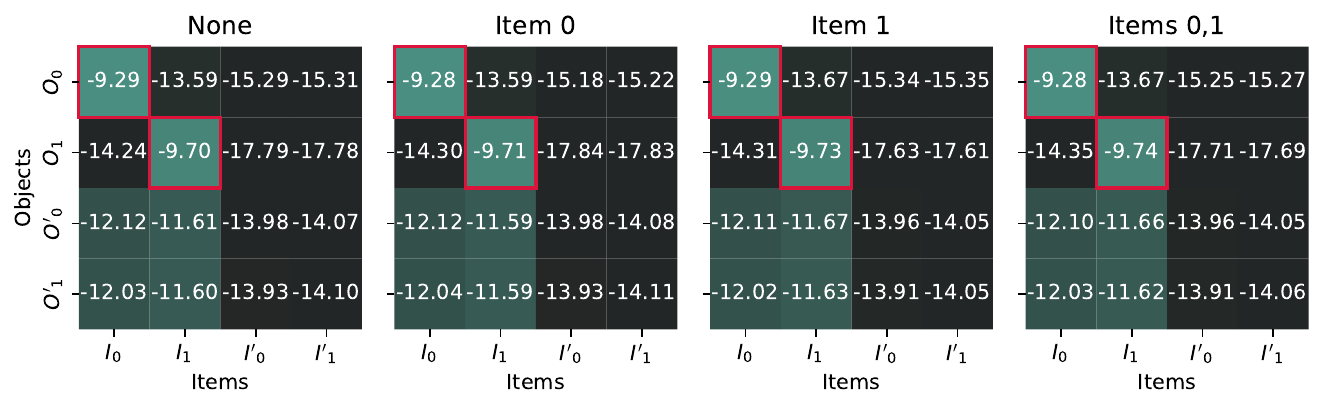}\vspace{-0.5em}
    \caption{Color activation replacements}
    \label{fig:fact_color}
\end{subfigure}
\vspace{-5mm}
\caption{\textbf{Factorizability results.}
Each row shows the model's mean log probabilities of an item contained in an object.
The first grid in each case shows results with unaltered activations.
Squares highlighted in red denote the expected predictions based on our hypothesis.
Model outputs match hypothesis suggesting a multimodal binding ID mechanism.}
\vspace{-4mm}
\label{fig:factorizability}
\end{figure}

\paragraph{Results.}
\cref{fig:factorizability} shows the mean log probability of choosing an item before and after interventions. We show the factorizability results for object patch tokens, color tokens and item tokens. 
In \cref{fig:fact_object}, the \textbf{first} grid shows the results when the \textit{activations are unaltered}. 
As expected, for objects $O_0$ and $O_1$, items $I_0$ and $I_1$ are chosen at a higher rate, respectively and for objects $O'_0$ and $O'_1$, items $I_0$ and $I_1$ are chosen at a roughly equal rate since these objects do not exist in the image. 
In the \textbf{second} grid, we replace $Z_{O_0}$ with $Z_{O'_0}$. Now, when the model is queried for the item contained by $O'_0$, the model picks item $I_0$ over $I_1$.
The \textbf{third} grid follows the same pattern, $Z_{O_1}$ is replaced by $Z_{O'_1}$ resulting in $O'_1$ containing $I_1$.
Finally, both object activations are replaced in the \textbf{fourth} grid and we observe that the model chooses $I_{0/1}$ for $O'_{0/1}$ respectively.
Note that when the object patches are replaced,
the color of the new object no longer matches the color description in the text. Nevertheless, the new object is still associated with the same item as the original object, as they both contain the same binding vector.

We observe a similar behavior for replacing items in \cref{fig:fact_item}. When $Z_{I_k}$ is replaced by $Z_{I'_k}$, the model prefers item $I'_k$ for object $O_k$.
However, when we intervene on the color activations $Z_{C_k}$, the results are similar to when there are no interventions (\cref{fig:fact_color}). This is expected since both $Z_{C_k}$ and $Z_{C'_k}$ contain the same binding ID vectors.

\subsection{Position Independence} 
\label{subsubsec: position independence}
Next, we hypothesize that the associations formed by the model are invariant to the activation positions of the object, color, or item, as they rely solely on the binding IDs.
This implies that swapping the positions of $Z_{O_0}$ and $Z_{O_1}$ should not change items associated with the objects.
To validate this, we first obtain the activations of the context tokens $Z_c$ (\cref{subsubsec: factorizability}). Then, we compute a new stack of activations $Z_c^{*}$ wherein the positions of $Z_{O_0}$ and $Z_{O_1}$ are altered, following the procedure described in~\cite{feng2024how}, adapted for models that use Rotary Position Embedding (RoPE)~\cite{su2021roformer}. Unlike absolute position embeddings, RoPE incorporate positional information only through the attention score computations, without injecting it directly into the residual stream activations.

\begin{figure*}[t]
\centering
\begin{subfigure}[b]{.33\linewidth}
    \centering
    \includegraphics[width=\linewidth]{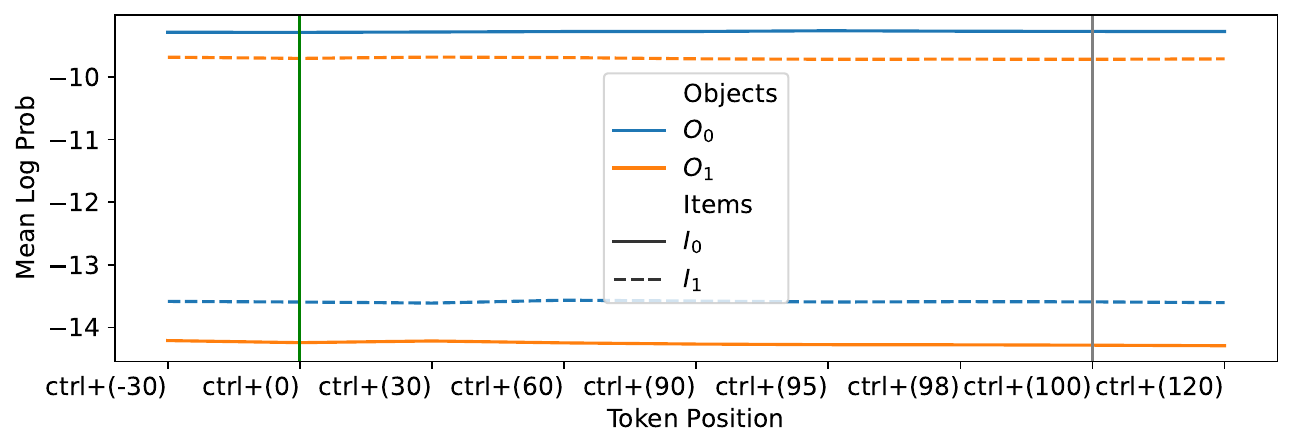}
    \caption{Object}
    \label{fig:pos_ind_object}
\end{subfigure}
\hfill
\begin{subfigure}[b]{.33\linewidth}
    \centering
    \includegraphics[width=\linewidth]{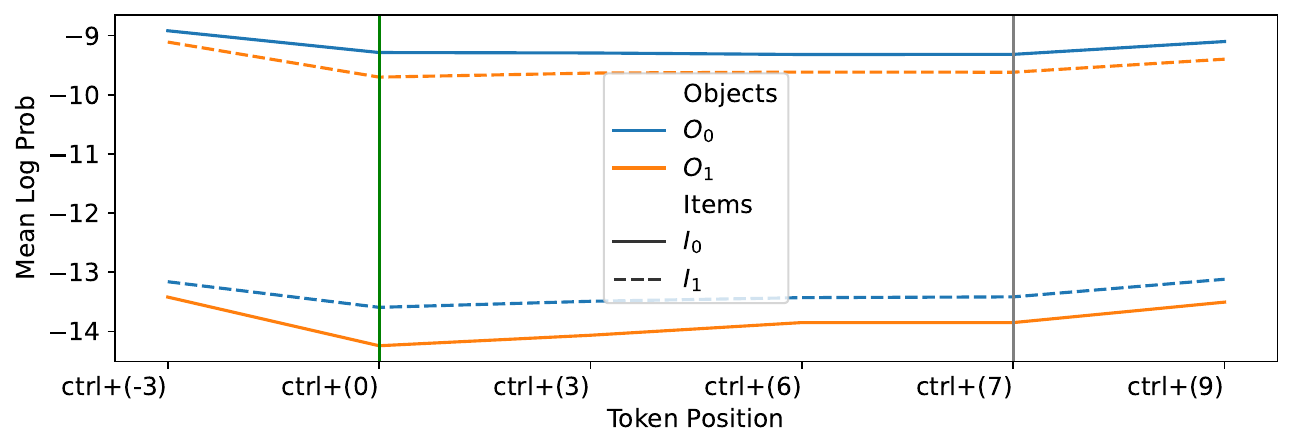}
    \caption{Item}
    \label{fig:pos_ind_item}
\end{subfigure}
\hfill
\begin{subfigure}[b]{.33\linewidth}
    \centering
    \includegraphics[width=\linewidth]{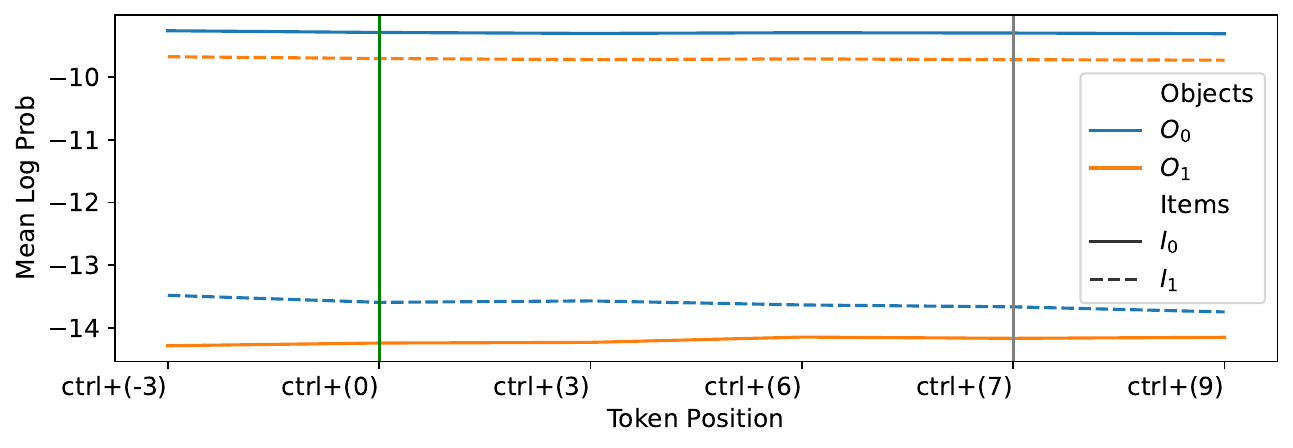}
    \caption{Color}
    \label{fig:pos_ind_color}
\end{subfigure}
\vspace{-6mm}
\caption{Position independence results. The integers in the x-axis show how much the position of the first and second objects/items/colors are incremented and decremented respectively. The green line corresponds to no change in positions and the gray line corresponds to swapped positions.
In all cases $O_k \leftrightarrow I_k$ (blue solid $O_0, I_0$ and oranged dashed $O_1, I_1$) have a higher probability than $O_k \leftrightarrow I_k'$.}
\vspace{-2mm}
\label{fig:pos_ind}
\end{figure*}

\paragraph{Results.}
\cref{fig:pos_ind} shows the mean log probabilities when the positions of $Z_{O_0}$ and $Z_{O_1}$ are progressively adjusted to get closer and ultimately swapped.
We observe that the model answers with the correct item regardless of positions.

\begin{figure}[t]
\centering
\includegraphics[width=0.45\textwidth]{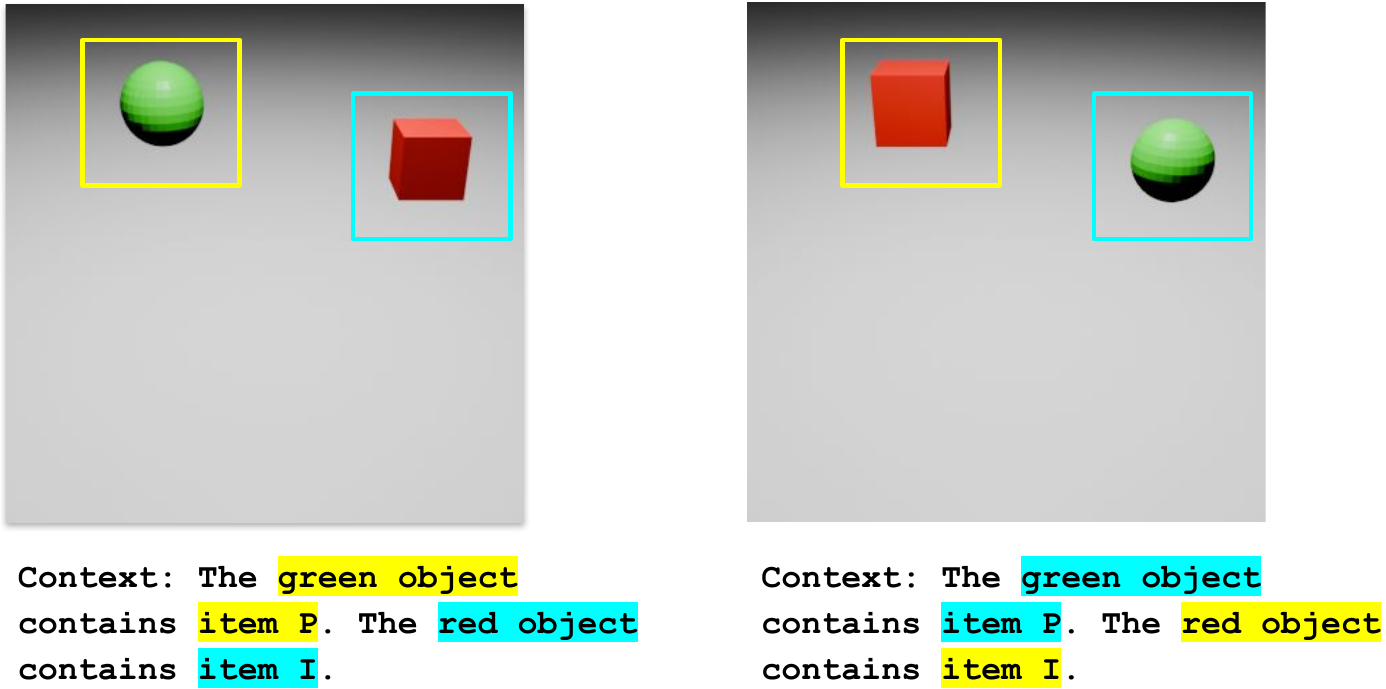}
\caption{Mean intervention samples.}
\label{fig:mean_ab_fig}
\vspace{-3mm}
\end{figure}

\subsection{Mean Interventions}
\label{subsubsec: mean interventions}

The factorizability and position independence results show that binding vectors are contained within the activations corresponding to the object, color, and item tokens and cause the model to form associations across image and text.
If binding vectors were directly accessible, we could interchange them to observe if the model changes its answer.
While this is not feasible, we can approximate the difference in binding vectors from the difference in activations.
To estimate $\Delta_O = b_O(1) - b_O(0)$, we consider two instances of the Shapes task as shown in \cref{fig:mean_ab_fig}. Let $O_0$, $O_1$ denote the objects in the first instance and $O'_0$, $O'_1$ denote the objects in the second instance. Notice that both $O_0$ and $O'_1$ are the same object, a \textit{green sphere}. However, we expect their activations to contain different binding IDs. We can now estimate $\Delta_O$ as the difference $Z_{O'_1} - Z_{O_0}$. Concretely, we compute $\Delta_O$ as the mean of the difference of activations over multiple pairs of instances ($\Delta_O \approx mean_{O_0, O'_1}[Z_{O'_1} - Z_{O_0}]$). Similarly, we compute $\Delta_C = b_C(1) - b_C(0)$ and $\Delta_I = b_I(1) - b_I(0)$ from the color and item activations.

Using these mean vectors ($\Delta_O$, $\Delta_C$, $\Delta_I$), we can now edit the binding vectors in the activations to alter the model response. For any new instance with the context $c^* = ctxt(O^*_0\leftrightarrow C^*_0\leftrightarrow I^*_0, O^*_1\leftrightarrow C^*_1\leftrightarrow I^*_1)$, we can alter the binding vector of the objects as ${Z_{O^*_0} := Z_{O^*_0} + \Delta_O}$ and ${Z_{O^*_1} := Z_{O^*_1} - \Delta_O}$. This should result in a swap of object-item binding with ${O^*_0}$ and ${O^*_1}$ being bound to ${I^*_1}$ and ${I^*_0}$ respectively. Similarly, altering the binding vector of the items as $Z_{I^*_0} := Z_{I^*_0} + \Delta_I$ and $Z_{I^*_1} := Z_{I^*_1} - \Delta_I$ should also exchange the model response. Altering the binding vectors in color token activations will make the model now associate $O_k$, $C_{k'}$ and $I_k$ where $k \neq k'$. However, $O_k$ is still bound to $I_k$, and we expect no change in response.

\paragraph{Results.}
\cref{tab:mean_ab} shows the accuracy, measured as the fraction of samples where the correct item has the highest log probability among the possible items in the context. As expected, both object and item interventions individually change the model's response, while color interventions do not.
Further, simultaneously performing object and item interventions restores the model's original response since they now have the same binding IDs.
We also repeat these experiments with random vectors that have the same magnitude but different directions. These vectors do not alter the model response, indicating that the specific directions of the mean vectors causally affect the binding.

\begin{table}[t]
\centering
\small
\begin{tabular}{ccccc}
\toprule
\multirow{2}{*}{Condition}  & \multicolumn{2}{c}{Mean vectors} & \multicolumn{2}{c}{Random vectors} \\
&  $O_0 \leftrightarrow I_0$ & $O_1 \leftrightarrow I_1$ & $O_0 \leftrightarrow I_0$ & $O_1 \leftrightarrow I_1$\\
\midrule
None    & 1.00 & 1.00 & - & - \\
O       & 0.00 & 0.05 & 1.00 & 1.00 \\
I       & 0.05 & 0.00 & 1.00 & 1.00 \\
C       & 1.00 & 1.00 & 1.00 & 1.00 \\
O, I    & 1.00 & 0.95 & 1.00 & 1.00 \\
O, I, C & 1.00 & 0.95 & 1.00 & 1.00 \\
\bottomrule
\end{tabular}
\vspace{-2mm}
\caption{Mean ablation accuracies: Object (O), Item (I), Color (C).}
\label{tab:mean_ab}
\vspace{-4mm}
\end{table}

\subsection{Experimental Details}
Throughout the paper, we report results with LLaVA-OneVision-7B~\cite{li2024llava}, which uses the SigLIP~\cite{zhai2023sigmoid} vision encoder and encodes multiple crops from a single image. The Shapes task images are of size 384$\times$384, with each object appearing in two crops and occupying 5$\times$5 patch tokens. 
Empirically, we found that when intervening on object token activations, a 3-token padding on all sides in both crops yields optimal results.
To estimate the difference of binding vectors, we use a separate set that contains different shapes (frustum, pyramid, prism and toroid), colors (lime, pink, gold, brown, orange and azure) and items (lowercase English alphabet). All colors and items span two text tokens.

\section{Related Work}
\label{sec: related work}

The Binding ID mechanism explains how LLMs associate concepts in context, leading to the identification of a binding subspace where bound tokens have a higher similarity than unbound ones~\cite{feng2025monitoring}. Concurrently, researchers uncovered circuits for entity tracking in LLMs, allowing inference of entity properties from context~\cite{prakashfine}. The Shapes task is inspired by the text-based entity tracking task~\cite{kim2023entity}, which requires predicting an entity's state based on its initial description and applied operations.

Prior works have analyzed attention heads in VLMs to understand visual processing \cite{kaduri2024s}, shown that object information is localized to corresponding image token positions \cite{neo2025towards}, and developed methods to manipulate image token representations to mitigate hallucinations \citep{jiang2025interpreting}. Our work complements these efforts by examining the association between image and text representations.

Benchmarks like VTQA~\cite{chen2024vtqa} and MuMuQA~\cite{reddy2022mumuqa} pose multi-hop questions that require synthesis of visual and textual information, going beyond traditional VQA where answers rely primarily on visual inputs. They present an opportunity to explore how mechanisms such as Binding IDs could enhance reasoning in complex, realistic scenarios.

\section{Conclusion}
\label{sec:conclusion}

In this work, we explore how in-context associations occur in VLMs. We formulate the Shapes task, a simple and controlled QA task which requires the model to associate 3D objects in an image with their references in the text. Through experiments, we demonstrate that VLMs utilize binding ID vectors to bind concepts across image and text.

\balance
{
\small
\bibliographystyle{ieeenat_fullname}
\bibliography{bib/longstrings, bib/main}
}


\end{document}